\documentclass{article} 
\usepackage{iclr2025_conference,times}

\usepackage{amsmath,amsfonts,bm}

\def\eqref#1{equation~\ref{#1}}

\def\1{\bm{1}}

\DeclareMathAlphabet{\mathsfit}{\encodingdefault}{\sfdefault}{m}{sl}
\SetMathAlphabet{\mathsfit}{bold}{\encodingdefault}{\sfdefault}{bx}{n}

\usepackage{hyperref}
\usepackage{algorithm}
\usepackage{algpseudocode}
\usepackage{url}
\usepackage{booktabs}
\usepackage{float}
\usepackage{subfigure}
\usepackage{graphicx}
\usepackage{caption}
\usepackage{pifont}
\usepackage[capitalise]{cleveref}

\title{SMiR: Efficient \textbf{S}ynthetic Data Pipeline to Improve \textbf{M}ulti-\textbf{i}mage \textbf{R}easoning}

\author{%
  \textbf{Andrew Li$^{1}$}\thanks{Equal Contribution.} \,\thanks{Corresponding author: \texttt{andrewli2403@berkeley.edu}. Work done during internship at Together AI.} \quad \textbf{Rahul Thapa$^{2}$}\footnotemark[1] \quad \textbf{Rahul Chalamala$^{3}$} \quad \textbf{Qingyang Wu} \\
  \textbf{Kezhen Chen}
  \quad \textbf{James Zou$^{2}$} \\
  Together AI \quad $^1$University of California, Berkeley \quad $^2$Stanford University \quad $^3$Caltech \\
}

\iclrfinalcopy

\begin{document}

\maketitle

\begin{abstract}
Vision-Language Models (VLMs) excel at understanding single images, aided by high-quality instruction datasets. However, multi-image reasoning remains underexplored in the open-source community due to two key challenges: (1) scaling datasets with correlated images and complex reasoning instructions is resource-intensive, and (2) robust evaluation benchmarks for multi-image tasks are lacking. To address this, we introduce \textsc{SMiR}, a synthetic data-generation pipeline for multi-image reasoning, along with a high-quality dataset generated using this pipeline. \textsc{SMiR} efficiently extracts correlated images via multimodal embeddings, integrates visual and descriptive information, and leverages open-source LLMs to generate quality instructions. Using this approach, we produce 160K synthetic training samples, offering a cost-effective alternative to closed-source solutions. Additionally, we present \textsc{SMiR-Bench}, a multi-image reasoning benchmark comprising 200 diverse examples across seven complex reasoning tasks. \textsc{SMiR-Bench} is multi-turn and employs a VLM judge to evaluate free-form responses, providing a comprehensive assessment of model expressiveness and reasoning capability across modalities. We demonstrate the effectiveness of \textsc{SMiR} by fine-tuning open-source VLMs and evaluating them on \textsc{SMiR-Bench}. Our codebase is available at \url{https://github.com/togethercomputer/SMiR}.



\end{abstract}



\section{Introduction}

Vision-Language Models (VLMs) have demonstrated strong performance on single-image tasks, particularly open-source models trained with high-quality instruction datasets \citep{laurenccon2024matters, zhang2023instruction, xu2022multiinstruct}. However, open-source VLMs struggle with multi-image reasoning tasks—such as comparing and analyzing relationships between multiple images—lagging behind closed-source models like GPT-4 \citep{achiam2023gpt}, Claude 3.5 Sonnet \citep{anthropic2024claude3.5addendum}, and Gemini 1.5 \citep{reid2024gemini}. This performance gap likely stems from the superior multi-image datasets available to closed-source models, which are essential for effective reasoning across multiple images.

A key challenge in multi-image reasoning is collecting and curating large-scale datasets with highly correlated images. Existing datasets such as MANTIS \citep{jiang2024mantis} often pair unrelated images, simplifying the task and limiting the complexity of reasoning. MMDU-45K \citep{liu2024mmdu} improves upon this by clustering images based on captions and category tags but overlooks purely visual relationships. Consequently, datasets that explicitly ensure image correlation are needed to advance complex multi-image reasoning.

Scaling such datasets further exacerbates the challenge. Human curation, as required in datasets like MultiInstruct \citep{xu2022multiinstruct}, is resource-intensive. While synthetic datasets generated via GPT-4 \citep{peng2023instruction, wang2023see} offer a scalable alternative \citep{liu2024mminstruct, li2024multimodal, li2023mimic}, they remain expensive and difficult to scale effectively.

Another challenge is evaluation. Existing multi-image benchmarks, such as MMIU \citep{meng2024mmiu} and MMMU \citep{yue2024mmmu}, rely on multiple-choice questions, failing to assess the depth of a model’s reasoning. While some benchmarks incorporate free-form responses \citep{liu2024mibench}, they lack multi-turn interactions, which are essential for evaluating complex reasoning processes.

To address these issues, we introduce \textsc{SMiR}, a synthetic data generation pipeline for multi-image reasoning, and \textsc{SMiR-Bench}, a multi-image evaluation benchmark. Our key contributions are:

\begin{itemize}
    \item We propose \textsc{SMiR}, a pipeline that efficiently identifies correlated images using multimodal embeddings, applying cluster sampling and graph iteration sampling to ensure diversity and robustness in instruction tuning.
    
    \item We develop a scalable synthetic data generation pipeline using open-source LLMs such as Meta Llama 3.1 70B Instruct Turbo \citep{dubey2024llama}, reducing costs by up to 50x and increasing speed by up to 10x compared to closed-source alternatives \citep{kirkovska2024llama}.
    
    \item We introduce \textsc{SMiR-Bench}, a multi-turn evaluation benchmark that assesses models via free-form responses and reasoning justifications, using GPT-4-Turbo and other baselines for comparison. Models trained on \textsc{SMiR} improve win rates by up to 8\% over existing instruction-tuned models.
\end{itemize}

By advancing both dataset generation and evaluation, \textsc{SMiR} facilitates the development of stronger open-source VLMs for multi-image reasoning.



\newcommand{\cmark}{\ding{51}}
\newcommand{\xmark}{\ding{55}}

\begin{table}[!h]
    \caption{Comparison of datasets highlighting key characteristics and methodologies.}
    \centering
    \resizebox{\textwidth}{!}{%
    \begin{tabular}{l|cccc}
    \toprule
       Datasets  & Multimodal Embedding & Correlated Images & No Human-Annotation & Open-Source LLM\\
    \midrule 
       Mantis \cite{jiang2024mantis} & \xmark & \xmark & \xmark & \xmark\\
       MMDU \cite{liu2024mmdu} & \xmark & \cmark & \xmark & \xmark\\
       SMiR   & \cmark & \cmark & \cmark & \cmark\\
    \bottomrule
    \end{tabular}
    }
    \label{tab:dataset_comparison}
\end{table}

\begin{figure}[!h]
    \centering
    \includegraphics[width=0.9\linewidth]{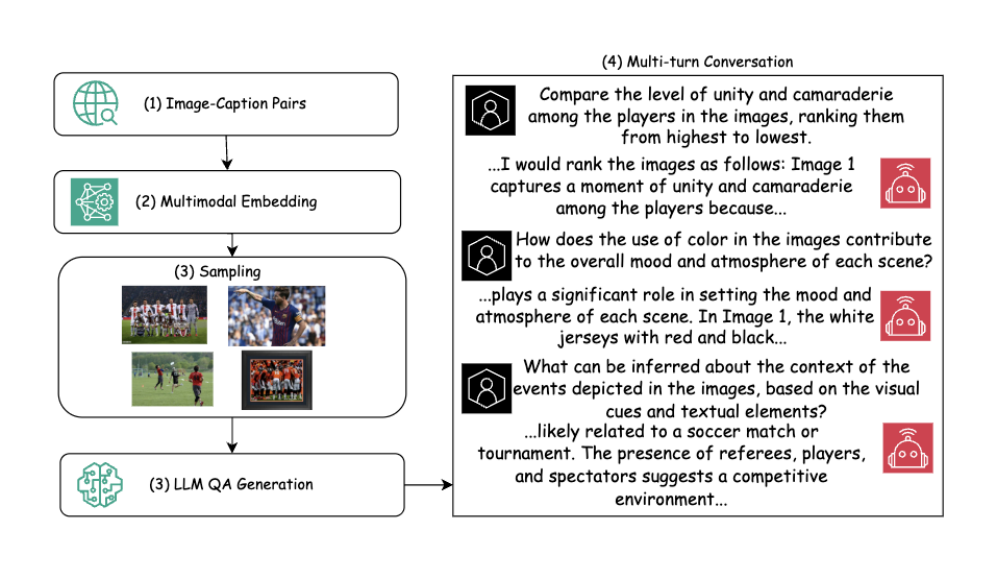}
    \caption{Our end-to-end pipeline converts image-caption pairs into synthetic multi-turn conversations using multimodal embeddings, strategic sampling, and LLM prompting. The example, based on a sports scenario, illustrates how the pipeline generates contextually rich dialogues by leveraging visual relationships.}

    \label{fig:pipeline}
\end{figure}



\begin{figure}[!h]
    \centering
    \includegraphics[width=0.9\textwidth]{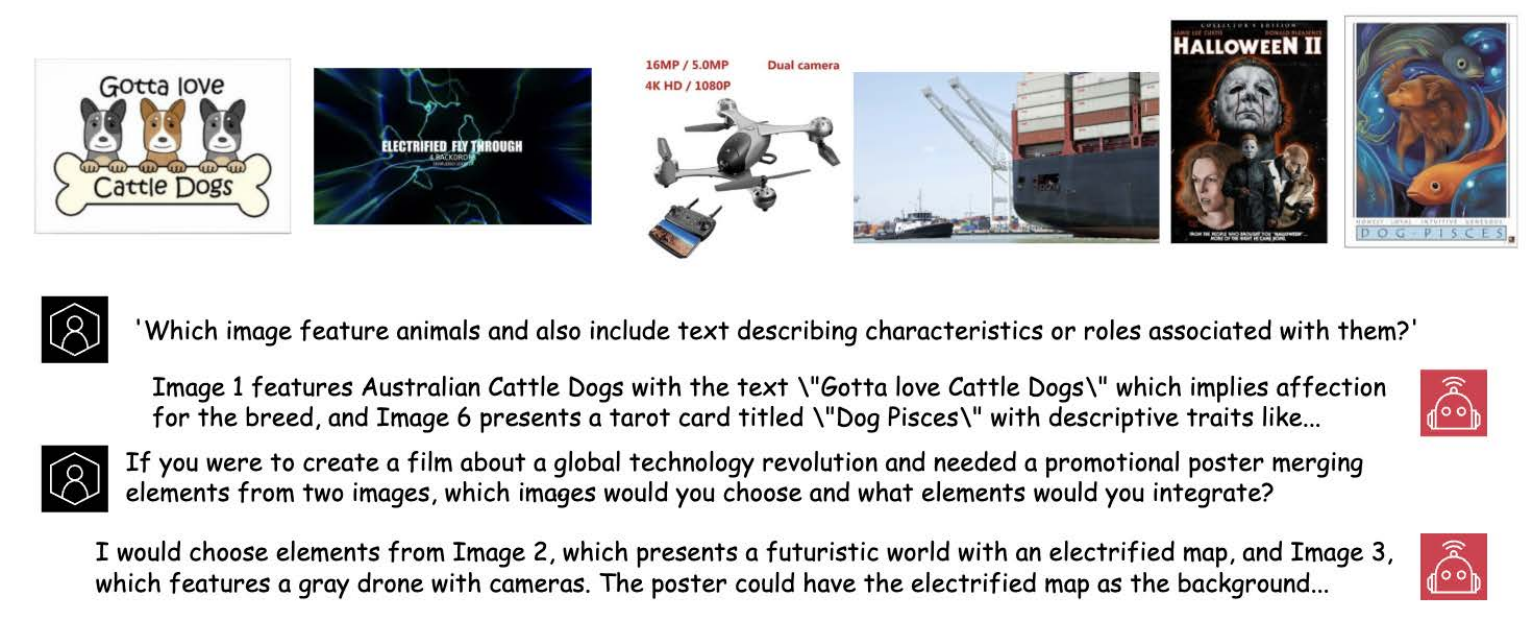}
    \vspace{0.5cm}
    \includegraphics[width=0.9\textwidth]{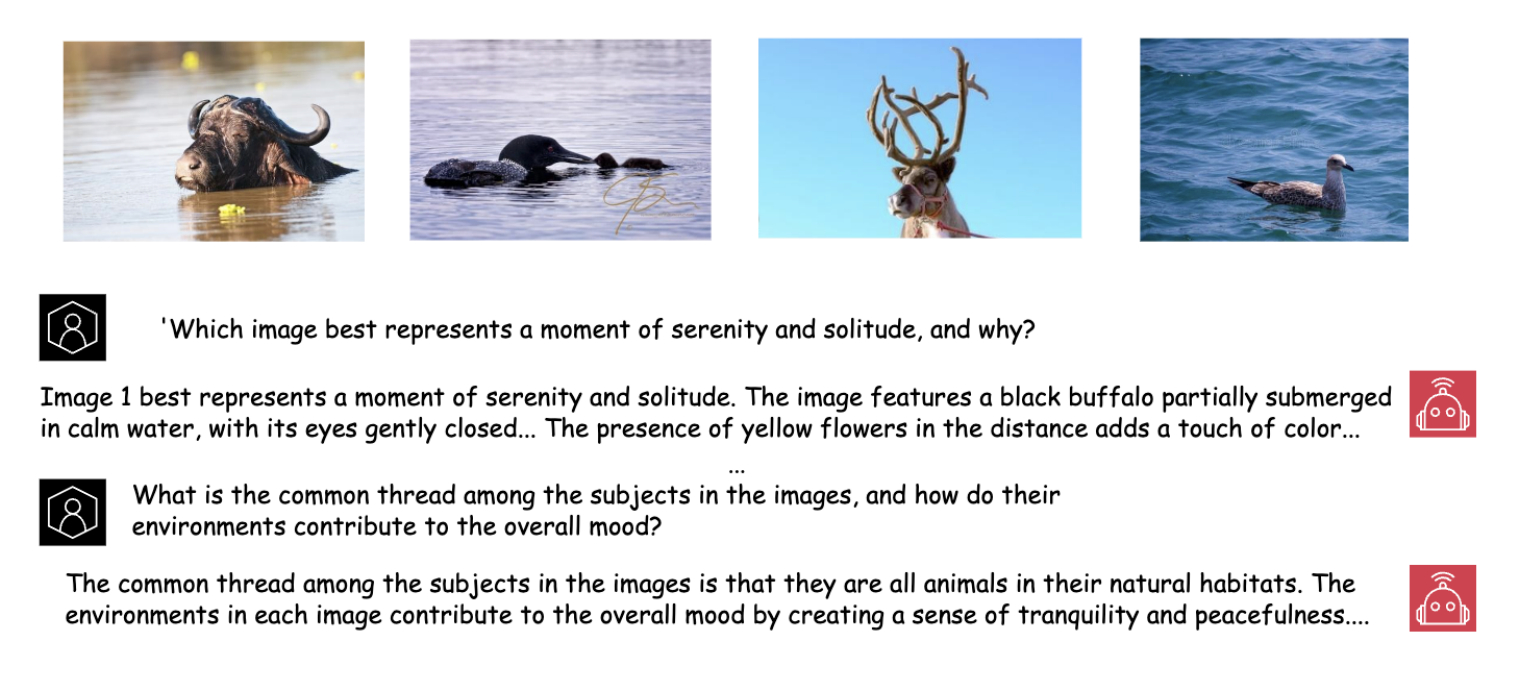}
    \caption{Comparison of an example from the MANTIS dataset, which includes samples from ShareGPT4V-PT where unrelated images such as animals and other objects are concatenated (top), vs. an example from the \textsc{SMiR} dataset, where only related images are grouped together (bottom), for multi-image reasoning.}

    \label{fig:comparison}
\end{figure}

\section{SMiR: Synthetic Multi-Image Reasoning Data Pipeline}

Starting with a large-scale image-caption dataset $D$ containing $N$ image-caption pairs $(I_i, C_i)_{i=1}^{N}$, \textsc{SMiR} first constructs multimodal embeddings for each pair. Then, sampling algorithms identify correlations between these embeddings. Finally, open-source LLMs are used to generate complex question-answer pairs based on the sampled correlations. The end-to-end pipeline is shown in \cref{fig:pipeline}.

\subsection{Multimodal Embedding Construction}

Embedding-based approaches have been widely used to model relationships between images and text. However, relying solely on image or caption embeddings limits the ability to fully capture the connections between visual content and descriptive text. A more comprehensive approach requires integrating both modalities.

To achieve this, we construct a multimodal embedding by combining SigLIP or CLIP image embeddings with their corresponding caption embeddings:

\begin{equation} 
E_{\text{multimodal}} = E_{\text{image}} + c \cdot E_{\text{caption}}
\end{equation}

where $E_{\text{multimodal}}$ represents the fused embedding, and $c$ adjusts the weight of the caption embedding. This parameter is tuned empirically based on small-scale human evaluations. For the ShareGPT4V dataset \citep{chen2023sharegpt4v}, $c=0.2$ provided the best balance between visual and textual contributions. The optimal value may vary depending on dataset quality and caption descriptiveness.

This multimodal approach enables better identification and clustering of correlated images by capturing both visual and textual relationships. We apply UMAP \citep{mcinnes2018umap} for dimensionality reduction, preserving essential data structures while facilitating efficient analysis and visualization.

\subsection{Random Sampling with Iteration} 

After constructing multimodal embeddings, we apply a random sampling algorithm to select semantically related images, ensuring diversity while preserving coherence. Given a set of embeddings $X = \{x_1, ..., x_n\}$, the algorithm proceeds as follows:

\begin{enumerate}
    \item Randomly selects an initial embedding and initializes the selected set $S$.
    \item Iteratively selects embeddings until reaching $N$ samples, using the probability distribution:
    \begin{equation} 
    p(x_j) \propto \frac{1}{\sum_{u \in S} \|x_j - x_u\|^k + \epsilon}, \quad x_j \in X \setminus S
    \end{equation}
    \item Samples the next embedding based on this probability, favoring closer points.
\end{enumerate}

Here, $k$ (default: 12) controls the emphasis on proximity—higher values prioritize nearby points more strongly by amplifying distance differences. This approach balances semantic coherence and diversity, allowing for meaningful multi-image relationships while introducing some randomness.

For example, as shown in \cref{fig:RSI_ex} (\cref{subsec:RSI_data}), this method effectively clusters related animal images while avoiding the disconnected groupings seen in simpler sampling methods. The full algorithm and pseudocode are provided in \cref{alg:random_sample_iteration} (\cref{subsec:RSI}).

By assembling related images before prompting the LLM, we generate richer contextual inputs, enabling the model to produce more nuanced and relevant synthetic data for multi-image reasoning tasks.

\begin{table}[!h]
\caption{Statistics of the \textsc{SMiR} dataset}
\centering
\begin{tabular}{l|r}
\toprule
\textbf{Metric} & \textbf{Value} \\
\midrule
Number of Samples & 160,000 \\
Maximum Number of Turns & 24 \\
Minimum Number of Turns & 2 \\
Average Number of Turns & 9.65 \\
Average Number of Images & 4.65 \\
Average User Tokens & 25.51 \\
Average Assistant Tokens & 124.32 \\
Open-Source LLM & Meta Llama 3.1 70B Turbo \\
\bottomrule
\end{tabular}
\label{tab:dataset_stats}
\end{table}

\subsection{Implementation}

Our implementation of the \textsc{SMiR} dataset construction builds on the structured sampling of correlated image-caption pairs. After grouping related images, we generate synthetic data by incorporating these captions into system prompts for an open-source LLM. We use Meta Llama 3.1 70B Turbo due to its efficiency, as it is up to 50 times cheaper and 10 times faster than GPT-4 \citep{kirkovska2024llama}. This process produces complex, multi-turn conversations between the user and assistant, as shown in \cref{tab:dataset_stats}. Sampled images and corresponding generated questions are illustrated in \cref{fig:GCMA_ex} and \cref{fig:RSI_ex} (\cref{sec:data_samples}).

To capture different aspects of visual reasoning, we design two types of system prompts. The first type focuses on shorter visual questions, often requiring OCR-based understanding. The second type is designed for long-form reasoning, where multi-step inference and deeper contextual understanding are needed. These prompts are inspired by LLaVA \citep{liu2024visual} and Chain-of-Thought reasoning \citep{wei2022chain}, enabling the generation of multi-turn conversations that enhance dataset complexity. The detailed structures of these prompts are provided in \cref{subsec:prompt_llava} and \cref{subsec:prompt_long}.

For data sources, we use ShareGPT4V \citep{chen2023sharegpt4v}, an open-source dataset that integrates data from LLaVA-Instruct \citep{liu2024improved} and COCO \citep{lin2014microsoft}. Originally, ShareGPT4V contained 100,000 captions generated by GPT-4 Vision, later expanding to 1.2 million image-caption pairs using an improved captioning model while maintaining quality and scale.

To ensure diversity in synthetic data, we process data in batches. Each batch consists of 20,000 images and generates 5,000 synthetic conversations, with each conversation containing an average of 4.65 images. In total, we use 640,000 image-caption pairs to produce 160,000 synthetic conversations. We employ a vector sampling algorithm alongside long-form question prompts to ensure a diverse set of multi-step reasoning tasks. The modular design of our pipeline allows easy adaptation to new datasets and applications, making \textsc{SMiR} a scalable solution for multi-image reasoning.



\section{Multi-Image Benchmark}

We introduce a benchmark that challenges VLMs to reason over multiple related images, analyze relationships, and derive meaning from image sequences. Unlike standard multiple-choice tasks, our benchmark requires detailed explanations, assessing models' ability to perform complex multimodal reasoning. To enable robust comparisons, we employ an open-ended evaluation framework where a judge model ranks responses in a pairwise structure, highlighting subtle differences in reasoning capabilities.

\subsection{Benchmark Overview}

We introduce \textsc{SMiR-Bench}, a multi-turn benchmark consisting of 200 examples spanning seven diverse topics. These topics include:

\begin{itemize}
    \item Bird: identifying species and reasoning over distinguishing features
    \item Matching: pairing photos based on visual similarities
    \item Object Character Recognition (OCR): reading and reasoning about academic texts
    \item Pattern: recognizing visual patterns in structured tasks
    \item Ranking: ordering objects based on contextual preferences
    \item Storytelling: narrating events from image sequences, as shown in \cref{fig:storytelling}
    \item Visual: establishing meaningful connections between images
\end{itemize}

For each category, we carefully selected 4-5 images from internet sources. Storytelling examples were specifically drawn from Shot2Story \citep{han2023shot2story20k} videos to leverage their sequential nature. GPT-4 was prompted to generate multi-turn question-answer conversations for these images, which were subsequently reviewed and refined by human annotators to ensure coherence and meaningful progression across dialogue turns.

\subsection{Evaluation Methodology}

\textsc{SMiR-Bench} extends the Auto-Hard-Auto v0.1 \citep{li2024crowdsourced} framework to multimodal evaluation. The process begins with all models completing the benchmark tasks. One model is selected as the baseline, and a GPT-4o judge conducts pairwise comparisons between the baseline model's responses and those of other models. The judge evaluates responses based on helpfulness, relevance, and conciseness, assigning comparative rankings such as $A \gg B$ (A significantly better than B), $A > B$ (A better than B), or $A = B$ (A equal to B). The framework aggregates these pairwise comparisons and confidence levels to produce an overall ranking of all models. An example of this evaluation process is illustrated in \cref{fig:storytelling}.

\section{Results}

To evaluate the effectiveness of our \textsc{SMiR} dataset, we finetune popular open-source VLMs and assess their performance on \textsc{SMiR-Bench}. Our experiments show that models fine-tuned on \textsc{SMiR} achieve up to an 8\% improvement over their base versions, demonstrating the dataset's value for enhancing multi-image reasoning capabilities.

We begin with two pre-trained models: Mantis-8B-siglip-llama3-pretrained \citep{jiang2024mantis} and Idefics-8B \citep{laurenccon2024matters}. The first model integrates a Llama-3-8B backbone with a SIGLIP encoder and is pre-trained on CC3M \citep{gan2022vision}. The second employs a Mistral-7B-v0.1 backbone with a perceiver-based projection \citep{jaegle2021perceiver} and is pre-trained on OBELICS.

Our evaluation consists of multiple stages. First, both pre-trained models undergo instruction tuning on MANTIS-Instruct (721K examples), resulting in Mantis-8B-siglip-llama3 and Mantis-8B-Idefics2. We then fine-tune these same pre-trained models on our \textsc{SMiR} dataset, producing SMiR-8B-siglip-llama3 and SMiR-8B-Idefics2.

To assess model performance, we conduct pairwise comparisons using GPT-4o as a judge model against three baselines:

\begin{itemize}
    \item \textbf{GPT-4-Turbo}, to compare our models against a leading closed-source solution (\cref{tab:gpt_baseline}).
    \item \textbf{Mantis-8B-siglip-llama3}, to evaluate the improvements from \textsc{SMiR} fine-tuning (\cref{tab:siglip_baseline}).
    \item \textbf{Mantis-8B-Idefics2}, to assess the performance gains of the Idefics-based model trained on \textsc{SMiR} (\cref{tab:idefics_baseline}).
\end{itemize}

Despite being trained on a significantly smaller dataset, models fine-tuned on \textsc{SMiR} outperform their MANTIS-tuned counterparts across all evaluation scenarios, demonstrating the value of our dataset for improving multi-image reasoning. Moreover, the significant performance gap between open-source and closed-source models on \textsc{SMiR-Bench} highlights the benchmark’s difficulty. Unlike prior benchmarks, \textsc{SMiR-Bench} focuses on highly correlated images, multi-turn question answering, and open-ended reasoning, making it a more rigorous test of multimodal understanding.

\begin{table}
\centering
\setlength{\tabcolsep}{4pt}
\caption{\textsc{SMiR-Bench} scores with Mantis-8B-siglip-llama3 baseline.}
\begin{tabular}{l|cccc}
\toprule
\textbf{Model Name} & \textbf{\textsc{SMiR-Bench}} & \textbf{$\Delta$} & \textbf{95\% CI} & \textbf{Average Tokens} \\
\midrule
Claude-3-Opus-20240229 & 97.4 & -- & (-1.3, 1.0) & 321 \\
Claude-3-5-Sonnet-20240620 & 97.1 & -- & (-1.3, 1.2) & 362 \\
GPT-4-Turbo & 96.4 & -- & (-1.8, 1.0) & 359 \\
Gemini-1.5-Pro & 96.3 & -- & (-1.3, 1.3) & 361 \\
GPT-4o & 91.5 & -- & (-2.4, 1.8) & 316 \\
\midrule
\textbf{\textsc{SMiR-8B-siglip-llama3}} & \textbf{58.1} & \textbf{+8.1\%} & (-4.6, 5.3) & 156 \\
\textbf{Mantis-8B-siglip-llama3} & \textbf{50.0} & -- & (0.0, 0.0) & 146 \\
\midrule
LLaVA-v1.6-Mistral-7B-HF & 15.8 & -- & (-2.4, 3.1) & 317 \\
Mantis-8B-siglip-llama3-pretraind & 8.7 & -- & (-2.1, 2.1) & 207 \\
\bottomrule
\end{tabular}
\label{tab:siglip_baseline}
\end{table}

\begin{table}
\setlength{\tabcolsep}{4pt}
\caption{\textsc{SMiR-Bench} scores with Mantis-8B-Idefics2 baseline.}
\centering
\begin{tabular}{l|cccc}
\toprule
\textbf{Model Name} & \textbf{\textsc{SMiR-Bench}} & \textbf{$\Delta$} & \textbf{95\% CI} & \textbf{Average Tokens} \\
\midrule
Claude-3-Opus-20240229 & 98.1 & -- & (-1.2, 0.9) & 321 \\
Claude-3-5-Sonnet-20240620 & 98.1 & -- & (-1.1, 1.1) & 362 \\
Gemini-1.5-Pro & 97.5 & -- & (-1.1, 0.9) & 361 \\
GPT-4-Turbo & 95.9 & -- & (-1.5, 1.6) & 359 \\
GPT-4o & 94.5 & -- & (-1.6, 1.5) & 316 \\
\midrule
\textbf{\textsc{SMiR-8B-Idefics2}} & \textbf{58.0} & \textbf{+8.0\%} & (-5.3, 3.9) & 157 \\
\textbf{Mantis-8B-Idefics2} & \textbf{50.0} & -- & (0.0, 0.0) & 171 \\
\midrule
Idefics2-8B & 29.7 & -- & (-3.7, 4.4) & 118 \\
LLaVA-v1.6-Mistral-7B-HF & 16.7 & -- & (-3.4, 3.0) & 317 \\
\bottomrule
\end{tabular}
\label{tab:idefics_baseline}
\end{table}


\section{Conclusion}

We introduces \textsc{SMiR}, a synthetic data pipeline for enhancing multi-image reasoning in open-source VLMs. By leveraging multimodal embeddings and grouping algorithms, our approach generates high-quality instruction tuning data, leading to up to an 8\% improvement on \textsc{SMiR-Bench}.  

However, certain limitations must be acknowledged. Random sampling with iteration is computationally intensive due to the need to recalculate distance embeddings for each newly sampled image. The scalability of our synthetic data also requires further investigation. Additionally, while \textsc{SMiR-Bench} provides a rigorous evaluation, our approach should be tested across other existing benchmarks to assess its generalizability.

\bibliography{main}
\bibliographystyle{iclr2025_conference}

\newpage
\appendix

\section{Related Works}

\paragraph{Vision Language Models}

We focus on instruction tuning Vision-Language Models (VLMs) that utilize a pretrained Large Language Model (LLM) backbone because this approach is cost-effective and more accessible for the open-source community. Since the backbone responsible for language understanding is already trained, the overall training process becomes simpler and requires fewer resources. Our primary task involves aligning the vision encoder—typically architectures like Vision Transformer (ViT) \citep{dosovitskiy2020image}, SigLIP \citep{zhai2023sigmoid}, or CLIP \citep{radford2021learning}—with the LLM backbone. This alignment is facilitated through linear layers that connect the vision encoder to the backbone, enabling the integration of visual and textual information. For instance, BLIP-2 \citep{li2023blip} uses OPT \citep{zhang2022opt} and FLAN-T5 \citep{chung2022scaling} as backbones, MiniGPT-4 \citep{zhu2023minigpt} utilizes Vicuna \citep{vicuna2023}, and Qwen-2-VL \citep{wang2024qwen2} employs Qwen-2-1.5B \citep{yang2024qwen2} as the language backbone. In this paper, we focus on creating a high-quality multi-image reasoning dataset for instruction tuning instead of large-scale interleaved pretraining datasets like OBELICS \citep{laurenccon2024obelics}, MINT-1T \citep{awadalla2024mint}, and LAION-5B \citep{schuhmann2022laion}.

\paragraph{Multi-Image Reasoning Data}

Recent advancements in multi-image reasoning instruction tuning datasets include MANTIS \citep{jiang2024mantis} and MMDU-45K \citep{liu2024mmdu}, both aiming to improve reasoning capabilities in VLMs. However, these datasets have limitations in their approaches. MANTIS randomly concatenates single image pairs from LLaVA-665k \citep{liu2024improved}, which often results in uncorrelated images within multi-image scenarios, potentially undermining the complexity of reasoning tasks. MMDU-45K attempts to address this issue by utilizing sentence transformers \citep{reimers2019sentence} with description text and clustering techniques to group related images, but does not consider visual components. The dataset is then further enhanced, assisted by GPT-4 to generate comprehensive answers for the grouped images. Building upon these efforts, \textsc{SMiR} introduces a novel approach that leverages both vital visual and caption information to ensure highly correlated images within multi-image sets with the use of open-source LLMs. These scalable methods leads to the generation of more challenging questions that require deeper analysis and understanding of visual relationships, pushing the boundaries of multi-image reasoning capabilities in VLMs.

\paragraph{Multi-Image Reasoning Benchmarks}

Recent VLM benchmarks \citep{chiang2024chatbot, lin2024wildbench, liu2024convbench} have made strides by incorporating free-response evaluations, marking a significant improvement over traditional multiple-choice formats. However, these benchmarks still lack a comprehensive approach that combines automatic, multi-turn, and pairwise evaluation capabilities. Our benchmark addresses this gap, drawing inspiration from Auto-Hard-Auto v0.1 \citep{li2024crowdsourced}. We have adapted and expanded this framework to enable robust multimodal evaluation, providing a more holistic assessment of VLM performance across complex, multi-image reasoning tasks. This approach allows for a deeper analysis of both the final answers and the underlying reasoning processes employed by VLMs in real-world \textsc{SMiR-Bench} scenarios.

\begin{figure}
    \centering
    \includegraphics[width=0.9\linewidth]{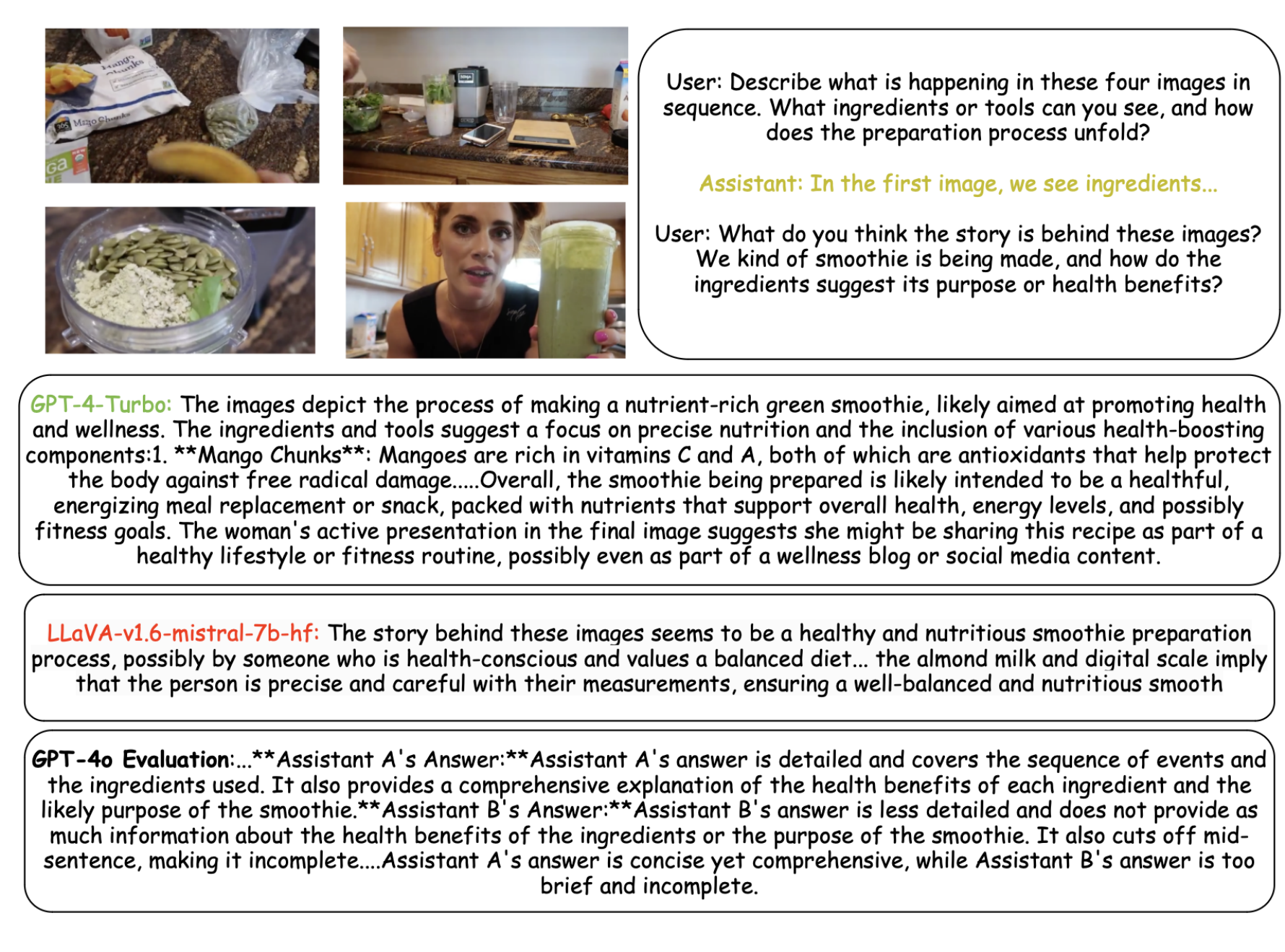}
    \caption{Evaluation Benchmark (Storytelling) Using GPT-4o as Judge}
    \label{fig:storytelling}
\end{figure}

\begin{table}
\centering
\setlength{\tabcolsep}{4pt}
\caption{\textsc{SMiR-Bench} scores with GPT-4-Turbo baseline.}
\begin{tabular}{l|cccc}
\toprule
\textbf{Model Name} & \textbf{\textsc{SMiR-Bench}} & \textbf{$\Delta$} & \textbf{95\% CI} & \textbf{Average Tokens} \\
\midrule
Claude-3.5-Sonnet-20240620 & 54.2 & -- & (-4.0, 4.3) & 362 \\
\textbf{GPT-4-Turbo} & \textbf{50.0} & -- & (0.0, 0.0) & 359 \\
Gemini-1.5-Pro & 49.5 & -- & (-4.3, 4.5) & 361 \\
GPT-4o & 46.3 & -- & (-4.0, 4.5) & 316 \\
Claude-3-Opus-20240229 & 34.4 & -- & (-4.6, 4.4) & 321 \\
\midrule
\textbf{\textsc{SMiR-8B-siglip-llama3}} & \textbf{5.0} & \textbf{+.8\%} & (-2.2, 2.7) & 183 \\
\textbf{Mantis-8B-siglip-llama3} & \textbf{4.2} & -- & (-1.2, 2.2) & 146 \\
\midrule
\textbf{\textsc{SMiR-8B-Idefics2}} & \textbf{4.5} & \textbf{+1.3\%} & (-1.3, 1.4) & 157 \\
\textbf{Mantis-8B-Idefics2} & \textbf{3.2} & -- & (-1.2, 1.2) & 171 \\
\midrule
Idefics2-8B & 2.5 & -- & (-1.0, 1.2) & 118 \\
Mantis-8B-siglip-llama3-pretrained & 1.3 & -- & (-.8, .7) & 207 \\
LLaVA-v1.6-mistral-7b-hf & 1.1 & -- & (-.7, .7) & 317 \\
\bottomrule
\end{tabular}
\label{tab:gpt_baseline}
\end{table}

\section{Algorithm Details}
\label{sec:algorithm}
\label{app:appendix}
\subsection{Greedy Cluster Matching Algorithm}
\label{subsec:GCMA}

We present the pseudocode for the Greedy Cluster Matching in \cref{alg:greedy_matching}.

Let $C_S = \{S_1, ..., S_m\}$ and $C_C = \{C_1, ..., C_n\}$ be cluster sets from SigLIP and CLIP embeddings respectively. The algorithm proceeds as follows:

1. Select the largest cluster from either set:
   $X_{max} = \arg\max_{X \in C_S \cup C_C} |X|$

2. If $X_{max} \in C_S$, find the best match in $C_C$:
   $Y_{best} = \arg\max_{C_j \in C_C} score(X_{max}, C_j)$
   
3. If $X_{max} \in C_C$, find the best match in $C_S$:
   $Y_{best} = \arg\max_{S_i \in C_S} score(X_{max}, S_i)$

Where the score function is defined as:
$$score(A, B) = \frac{|A \cap B|}{\frac{|A| + |B|}{2}}$$

This process is repeated, greedily selecting the largest remaining cluster and finding its best match, until all clusters are matched or one set is exhausted.

\begin{algorithm}

\caption{Greedy Cluster Matching Algorithm}\label{alg:greedy_matching}
\begin{algorithmic}[1]
\Require Two lists of clusters $c1$ and $c2$
\Ensure List of matched cluster pairs
\State $c1 \gets \text{sort}(c1, \text{key}=\text{len}, \text{reverse}=\text{True})$
\State $c2 \gets \text{sort}(c2, \text{key}=\text{len}, \text{reverse}=\text{True})$
\State $\text{matched\_pairs} \gets []$
\State $\text{num\_samples} \gets 0$
\While{$c1$ is not empty and $c2$ is not empty}
    \If{$\text{len}(c1[0]) \geq \text{len}(c2[0])$}
        \State $\text{larger\_cluster} \gets c1.\text{pop}(0)$
        \State $\text{smaller\_list} \gets c2$
    \Else
        \State $\text{larger\_cluster} \gets c2.\text{pop}(0)$
        \State $\text{smaller\_list} \gets c1$
    \EndIf
    \State $\text{best\_match} \gets \text{None}$
    \State $\text{best\_score} \gets -1$
    \For{$i, \text{cluster}$ in $\text{enumerate}(\text{smaller\_list})$}
        \State $\text{overlap} \gets \text{len}(\text{set}(\text{larger\_cluster}) \cap \text{set}(\text{cluster}))$
        \State $\text{avg\_size} \gets (\text{len}(\text{larger\_cluster}) + \text{len}(\text{cluster})) / 2$
        \State $\text{score} \gets \text{overlap} / \text{avg\_size}$
        \If{$\text{score} > \text{best\_score}$}
            \State $\text{best\_score} \gets \text{score}$
            \State $\text{best\_match} \gets (i, \text{cluster})$
        \EndIf
    \EndFor
    \If{$\text{best\_match}$ is not None}
        \State $\text{best\_index}, \text{best\_cluster} \gets \text{best\_match}$
        \State $\text{union} \gets \text{list}(\text{set}(\text{larger\_cluster}) \cup \text{set}(\text{best\_cluster}))$
        \State $\text{matched\_pairs}.\text{append}(\text{union})$
        \State $\text{num\_samples} \gets \text{num\_samples} + \text{len}(\text{union})$
        \State $\text{smaller\_list}.\text{remove}(\text{best\_cluster})$
    \EndIf
\EndWhile
\State \Return $\text{matched\_pairs}, \text{num\_samples}$
\end{algorithmic}
\end{algorithm}

\newpage

\subsection{Random Sampling with Iteration}
\label{subsec:RSI}

\begin{figure}[t]
    \centering
    \includegraphics[height=8cm]{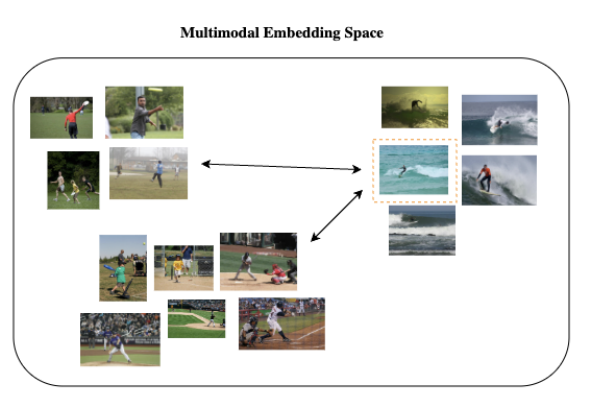}
    \caption{Sampling based on distances
    between multimodal embeddings of image-caption pairs. }
    \label{fig:sampling}
\end{figure}

We present the pseudocode for the Random Sampling with Iteration in \cref{alg:random_sample_iteration}.

Let $X = \{x_1, ..., x_n\}$ be the set of embeddings.

$k$ is a parameter that determines the power of the distance calculation (default to 12), and $N$ is the desired number of selected embeddings.

1. Randomly select an initial embedding: $s_1 \in X$

2. Initialize selected set $S = \{s_1\}$

3. For $i = 2$ to $N$:
$s_i = \arg\max_{x_j \in X \setminus S} \sum_{u \in S} \|x_j - x_u\|^k$,
$S = S \cup \{s_i\}$

4. Return $S$

This formulation captures the process of iteratively selecting embeddings based on their cumulative distance from all previously selected embeddings, raised to the power $k$.

\begin{algorithm}
\caption{Random Sampling with Iteration}\label{alg:random_sample_iteration}
\begin{algorithmic}[1]
\Require 
    \State $X$: Set of embeddings
    \State $N$: Number of samples to select
    \State $k$: Power factor for distance calculation (default: 12)
\Ensure Set of selected indices
\State $selected \gets []$
\State $n \gets |X|$ \Comment{Number of embeddings}
\For{$i = 1$ to $N$}
    \State $distances \gets \text{zeros}(n)$
    \If{$selected$ is empty}
        \State $sampled\_index \gets \text{random\_integer}(0, n-1)$
    \Else
        \For{$j = 0$ to $n-1$}
            \If{$j \in selected$}
                \State $distances[j] \gets \infty$
            \Else
                \State $distances[j] \gets \sum_{u \in selected} \|\|X[j] - X[u]\|\|^k$
            \EndIf
        \EndFor
        \State $inverted\_distances \gets \frac{1}{distances + \epsilon}$ \Comment{$\epsilon$ is a small constant}
        \State $distribution \gets \frac{inverted\_distances}{\sum inverted\_distances}$
        \State $sampled\_index \gets \text{random\_choice}(\text{range}(n), p=distribution)$
    \EndIf
    \State $selected.\text{append}(sampled\_index)$
\EndFor
\State \Return $selected$
\end{algorithmic}
\end{algorithm}

\section{Data Samples}
\label{sec:data_samples}

For the sake of brevity, we have included only two examples from the multi-image data we have generated, featuring questions with a related theme from both the Greedy Cluster Matching Algorithm and Random Sampling with Iteration, highlighting how the algorithms differ in their approach to generating questions.

\subsection{Greedy Cluster Matching Algorithm}

We present two novel algorithms, Greedy Cluster Matching and Random Sampling with Iteration, designed to group correlated images prior to leveraging an open-source Large Language Model (LLM) for synthetic data generation in multi-image reasoning tasks. The emphasis on correlated images is crucial, as it facilitates challenging multi-image reasoning scenarios. These scenarios require the model to identify intricate relationships and differentiate between visually similar scenes, thus enhancing the complexity and realism of the reasoning process.

\paragraph{Greedy Cluster Matching}
After employing HDBSCAN \citep{malzer2020hybrid}, a density-based clustering algorithm, to group the SigLIP and CLIP multimodal embeddings into coherent clusters, we developed a greedy algorithm to establish meaningful relationships between the two embedding spaces. Given cluster sets $C_S = \{S_1, ..., S_m\}$ from SigLIP and $C_C = \{C_1, ..., C_n\}$ from CLIP embeddings, the algorithm:

1) Orders clusters by size in descending order

2) Iteratively selects the largest remaining cluster $X_{max} = \arg\max_{X \in C_S \cup C_C} |X|$

3) For the selected cluster, finds its best match $Y_{best}$ from the other embedding space using:
   $Y_{best} = \arg\max_{Y} \frac{|X_{max} \cap Y|}{\frac{|X_{max}| + |Y|}{2}}$

This process continues until all clusters are matched or one set is exhausted, demonstrating high efficacy in producing quality image-caption pairs through dual-embedding validation. The score function, which measures overlap normalized by average cluster size, ensures each cluster corresponds to a semantically similar cluster from the other model. The detailed steps of this algorithm and pseudocode are presented in \cref{alg:greedy_matching} (\cref{subsec:GCMA}). 

\begin{figure}[H]
    \centering
    \includegraphics[width=.8\linewidth]{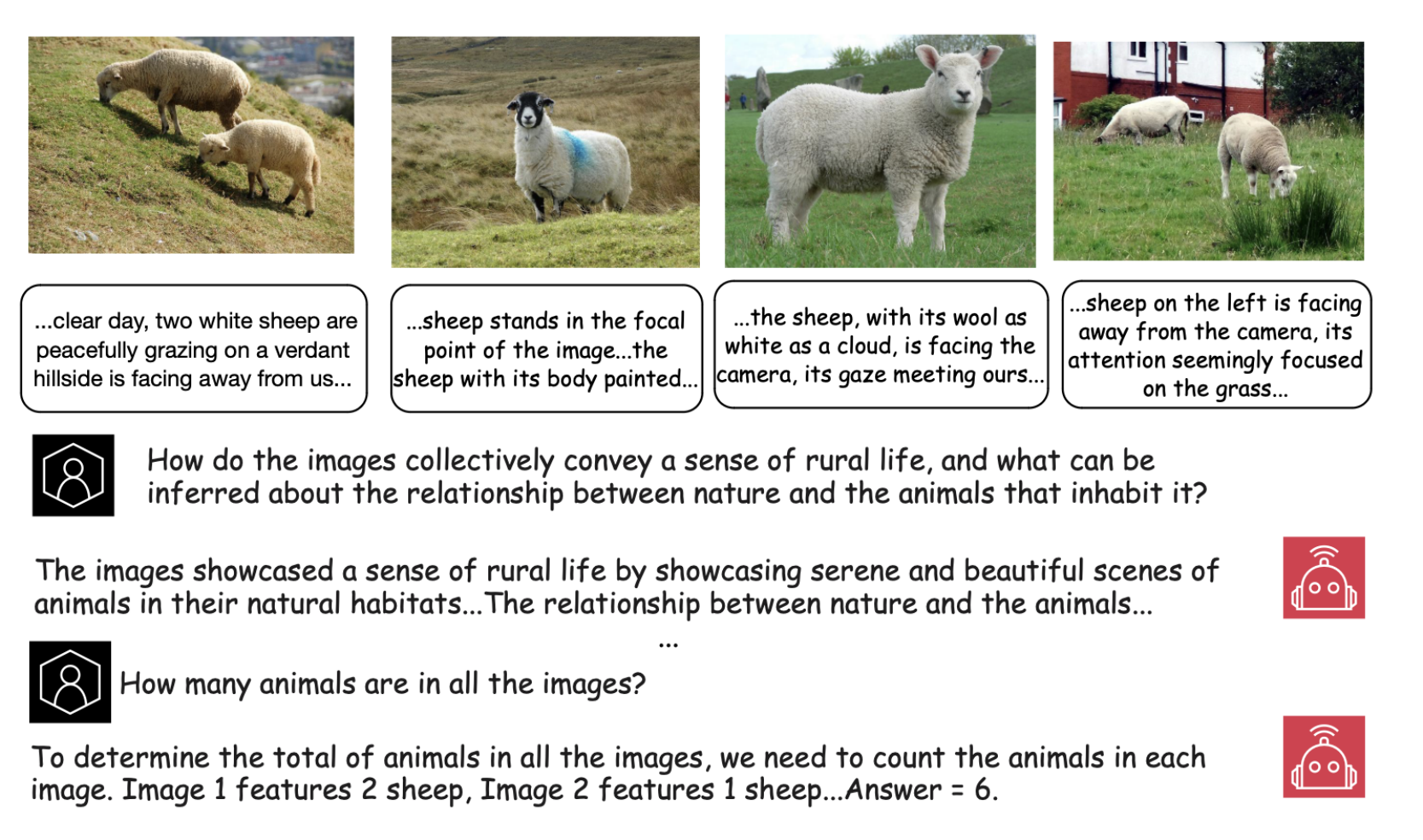}
    \caption{Images sampled from the same matched cluster often feature similar subjects or scenes.}
    \label{fig:GCMA_ex}
\end{figure}

While this approach effectively leverages both embedding models to confirm spatial relationships and associated semantic meanings, with images matched within clusters being corroborated by two independent embedding models, it has a notable limitation: sampling confined to matched clusters can lead to overly specialized image subjects, such as clustering only sheep-related images rather than diverse animal scenes, as demonstrated in \cref{fig:GCMA_ex} (\cref{subsec:GCMA_data}). This specialization occurs because sampling is restricted to a single matched cluster, potentially limiting the diversity of selected images.

Samples obtained through Greedy Cluster Matching typically feature similar subjects and shot compositions, but when paired with carefully crafted prompts, these similarities can be leveraged to generate more challenging and nuanced questions.

\label{subsec:GCMA_data}

\subsection{Random Sampling with Iteration}
\label{subsec:RSI_data}

Random sampling tends to yield greater diversity in subjects compared to Greedy Clustering, striking a balance between variety and relatedness that can potentially lead to more robust and wide-ranging question sets.

\begin{figure}[H]
    \centering
    \includegraphics[width=.8\linewidth]{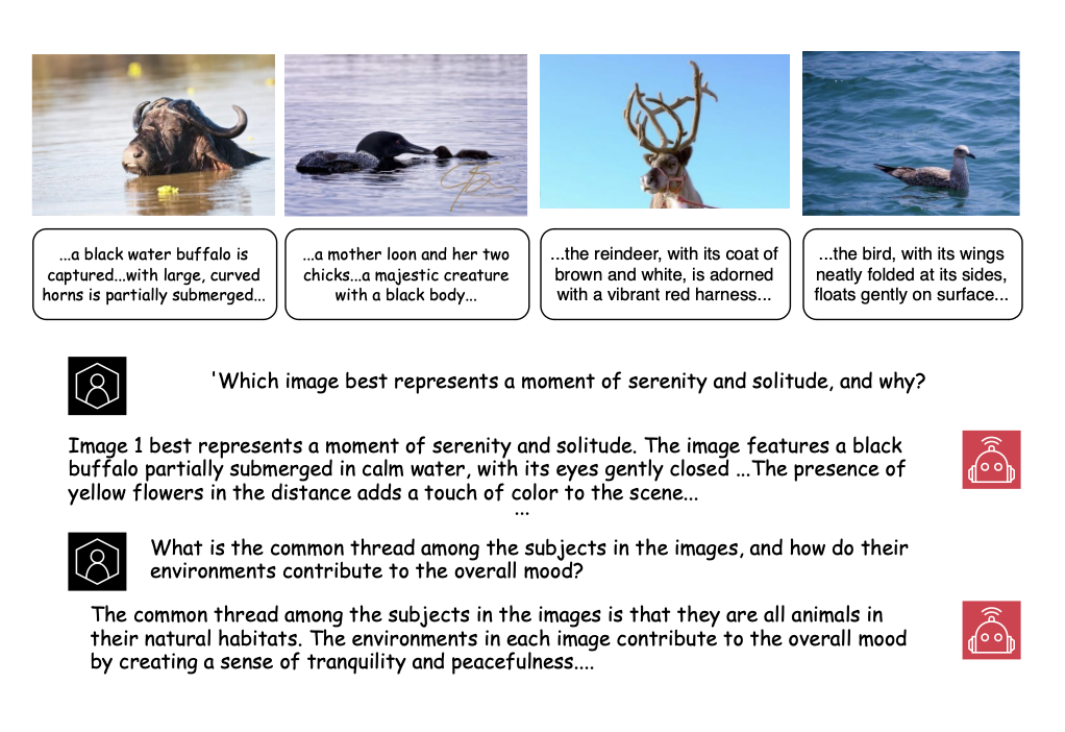}
    \caption{Images sampled using the iterative algorithm allow for different yet related subjects (e.g., various animal species)}
    \label{fig:RSI_ex}
\end{figure}

\newpage
\section{Prompt}
\label{sec:prompts}

While prompts play a crucial role in data generation, optimizing them remains a significant challenge. After numerous iterations, we identified two particularly effective prompts for multi-image data generation.

\subsection{LLaVA Style Prompt}
\label{subsec:prompt_llava}

Inspired by LLaVA \citep{liu2024visual}, our approach utilizes a specialized prompt to address simpler multi-image and single-image tasks, focusing on more straightforward visual comprehension and analysis.

{\fontfamily{cmtt}\selectfont You are an AI visual assistant that can analyze multiple images. You receive four to five images with their corresponding captions in an array. The task is to use the provided images and captions to create a challenging and complex question that involves comparison, ranking, storytelling, or logical reasoning across the images and then provide a detailed answer. The question should require a deep understanding of the visual content and advanced reasoning based on that content.

Create questions that ask to compare elements across the images, such as identifying which image best represents a critical or turning point moment, quality, or characteristic; formulate questions that require ranking the images based on intricate and plausible criteria, such as strategic importance, sequence, or visual impact; develop questions that involve piecing together a narrative from the images, understanding a sophisticated sequence of events, or explaining a complex progression shown; and ask questions that require advanced logical reasoning to deduce why certain elements are present, the purpose behind actions shown, or the broader implications of what is depicted.

Frame a question that requires advanced analyzing and reasoning about the images and their captions, and provide a detailed answer based on the visual content and captions, explaining the reasoning process and the conclusions drawn. Also, provide 3-4 follow-up questions to deepen the analysis based on the potential responses, and provide detailed answers for the follow-up questions as well.

Example questions include:
- ``Which image best represents the pivotal turning point of the event, and why?''
- ``Rank the images based on the strategic importance of the actions shown, from highest to lowest.''
- ``How do the images collectively tell the intricate story of the event, and what can be inferred about the key strategic moments?''
- ``What could be the underlying reasons behind the specific actions taken in each image, and how do they relate to the overall context in a broader sense?''

Example follow-up questions include:
- ``Why do you think the turning point identified in Image (image number) was critical to the outcome of the event?''
- ``Which image best represents the pivotal turning point of the event?\textbackslash n (A) Image 1\textbackslash n (B) Image 2\textbackslash n (C) Image 3\textbackslash n (D) Image 4\textbackslash n Answer with the letter only.''
- ``How many people are in all the images?\textbackslash nBegin with your reasoning, and then state your final count prefaced by \textbackslash``Answer:\textbackslash''''
- ``What additional details in Image (image number) support your interpretation of the story?''

Return your challenging and complex question and follow-up questions in the format:
``User: [INSERT QUESTION], Assistant: [INSERT ANSWER]
User: [INSERT FOLLOW-UP QUESTION 1], Assistant: [INSERT FOLLOW-UP ANSWER 1]
User: [INSERT FOLLOW-UP QUESTION 2], Assistant: [INSERT FOLLOW-UP ANSWER 2]
User: [INSERT FOLLOW-UP QUESTION 3], Assistant: [INSERT FOLLOW-UP ANSWER 3]...''
}

\subsection{Longer Prompt}
\label{subsec:prompt_long}

Our approach aims to generate more complex, multi-turn questions that require in-depth reasoning across multiple images.

{\fontfamily{cmtt}\selectfont You are an AI visual assistant capable of analyzing multiple images, including both visual content and textual elements using Optical Character Recognition (OCR). You will receive four to five images, each potentially accompanied by captions and containing text, numbers, signs, or other recognizable characters. Your task is to create a plausible and challenging question that involves comparison, ranking, storytelling, logical reasoning, or detailed textual analysis across the images, and then provide a detailed answer.

For Visual Analysis: Frame questions that require understanding and reasoning about the visual content, such as comparing elements across images, identifying which image best represents a specific moment, quality, or characteristic, ranking the images based on logical and plausible criteria (e.g., importance, sequence, or visual quality), or piecing together a narrative from the images to explain a sequence of events or the progression shown. Your questions should encourage deep engagement with the visual content and require advanced reasoning or interpretation.

For OCR-Based Analysis: Frame questions that require detailed analysis of the textual content in the images, such as counting specific items or words, identifying differences or similarities between the images, verifying the accuracy of information, or deducing logical conclusions based on the textual data. Your questions should encourage users to engage deeply with the textual content and require precise reasoning or interpretation.

Question Creation: Create a question that requires analyzing and reasoning about the images and their captions or textual elements. Provide a detailed answer based on the visual content, captions, and/or OCR analysis, explaining the reasoning process and conclusions drawn.

Follow-Up Questions: Additionally, include 3-4 follow-up questions that delve deeper into the analysis based on the initial question. Provide detailed answers for each follow-up question, further expanding on the reasoning and conclusions.

Example Questions:
- ``Which image best represents the climax of the event, and why?''
- ``Rank the images based on the level of engagement of the individuals shown, from highest to lowest.''
- ``How many times does the word \textbackslash``urgent\textbackslash'' appear across all images, and in which image is it most prominently displayed?''
- ``Identify whether all the images contain the same warning label. If there are differences, describe them.''

Example Follow-Up Questions:
- ``Why do you think the climax identified in Image (image number) was critical to the outcome of the event?''
- ``How do the textual differences in Image (image number) affect your understanding of the event's context?''
- ``What additional visual details in Image (image number) support your interpretation of the story?''
- ``How does the sequence of numbers in these images relate to the broader narrative depicted?''

Format: Return your challenging and complex question, detailed answer, and follow-up questions in the following format:
``User: [INSERT QUESTION], Assistant: [INSERT ANSWER]
User: [INSERT FOLLOW-UP QUESTION 1], Assistant: [INSERT FOLLOW-UP ANSWER 1]
User: [INSERT FOLLOW-UP QUESTION 2], Assistant: [INSERT FOLLOW-UP ANSWER 2]
User: [INSERT FOLLOW-UP QUESTION 3], Assistant: [INSERT FOLLOW-UP ANSWER 3]...''
}

\end{document}